# IMPLEMENTATION OF NLIZATION FRAMEWORK FOR VERBS, PRONOUNS AND DETERMINERS WITH EUGENE

Harinder Singh[1] and Parteek Kumar[2]

[1,2]Computer Science & Engineering Department Thapar University, Patiala, India
singh.harinder@outlook.com
parteek.bhatia@thapar.edu

## ABSTRACT

*UNL system is designed and implemented by a nonprofit organization, UNDL Foundation at Geneva in 1999. UNL applications are application softwares that allow end users to accomplish natural language tasks, such as translating, summarizing, retrieving or extracting information, etc. Two major web based application softwares are Interactive ANalyzer (IAN), which is a natural language analysis system. It represents natural language sentences as semantic networks in the UNL format. Other application software is dEep-to-sUrface GENErator (EUGENE), which is an open-source interactive NLizer. It generates natural language sentences out of semantic networks represented in the UNL format. In this paper, NLization framework with EUGENE is focused, while using UNL system for accomplishing the task of machine translation. In whole NLization process, EUGENE takes a UNL input and delivers an output in natural language without any human intervention. It is language-independent and has to be parametrized to the natural language input through a dictionary and a grammar, provided as separate interpretable files. In this paper, it is explained that how UNL input is syntactically and semantically analyzed with the UNL-NL T-Grammar for NLization of UNL sentences involving verbs, pronouns and determiners for Punjabi natural language.*

## 1. INTRODUCTION

Today all over the world work is in progress by various government/educational organizations and some individual researchers for technological development of most widely spoken natural languages. Machine Translation is most challenging task which is to be accomplished before excelling further in other sub domains of NLP (Natural Language Processing). The process of translating a text written in some language to another language's text is called Machine Translation.

There are many approaches used for machine translation. Interlingua based approach which is a rule based can be adopted for multilingual machine translation, as depicted in Figure 1.

In this approach, the target language is first transformed into some intermediate form, while preserving the semantic information associated with the words of source language. This intermediate form is independent of source language. From this intermediate form variety of natural languages can be generated, by a separate process using respective language word dictionaries and sets of grammar rules. If traditional statistical approach is followed then total of $n*(n\text{-}1)$ systems will be required for $n$ number of languages in multilingual machine translation. But, interlingua approach for machine translation system is a novel approach, here only $2n$ language systems are required, *i.e.*, two systems per language.

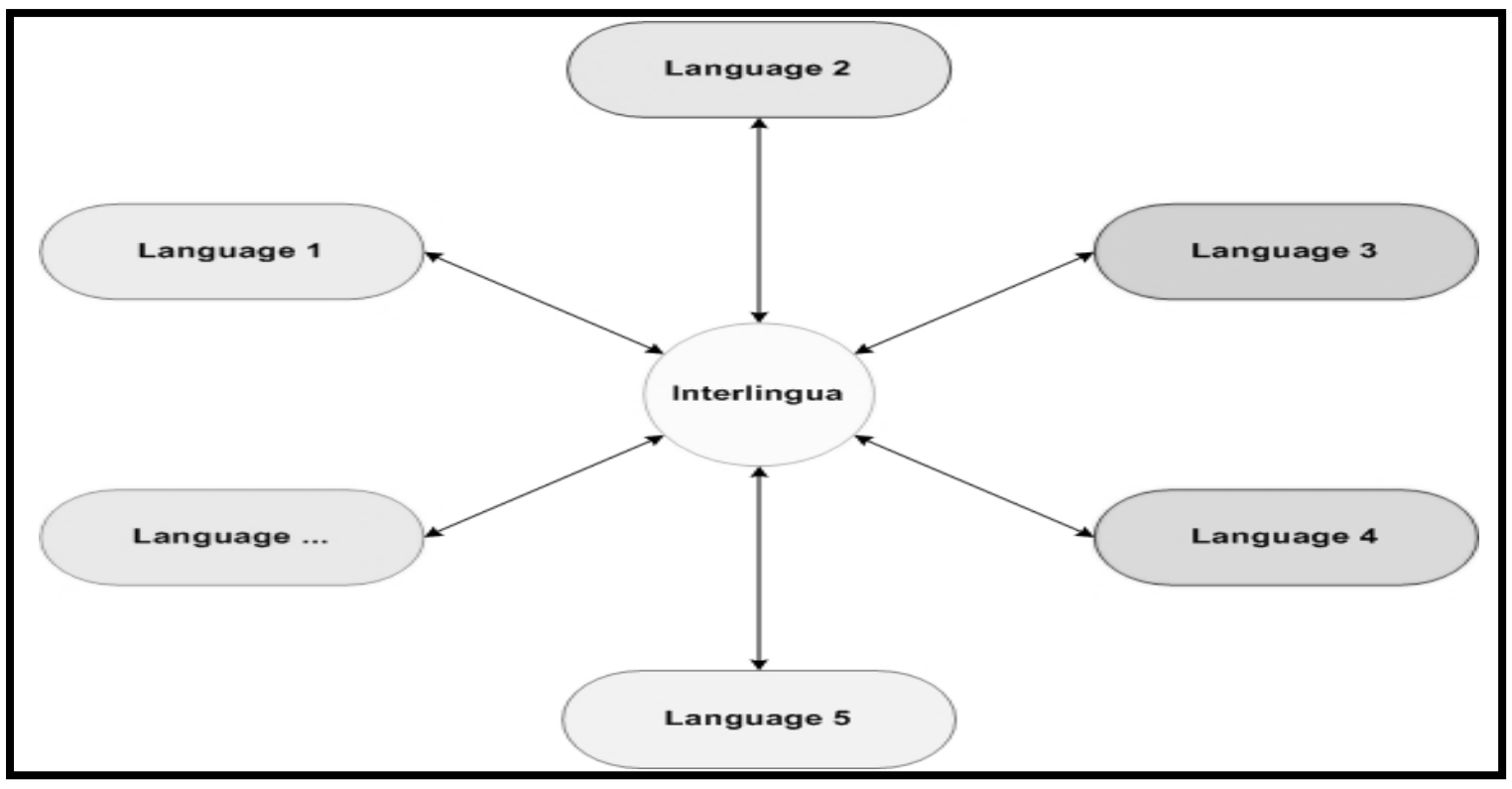

Figure 1. Model of Interlingual based Machine Translation

Universal Networking Language (*i.e.*, UNL) system is based on interlingua approach. UNL system is designed and implemented by UNDL foundation which is a non-profit organization based in Geneva, Switzerland in 1996. UNL system comprises of two components one is UNLizer for converting the source language to UNL form (an intermediate form) and other is NLizer for generating target language from UNL form, using respective language word dictionaries and sets of grammar rules. UNLization and NLization are different for different languages, because sentence structures of different languages are different.

In this paper, NLization for Punjabi language for verb, pronouns and determiners from UNL form (an intermediate form) is discussed. EUGENE (dEep-to-sUrface Natural language GENErator), is an open-source interactive NLizer, which is a new deconverter based on a high-level linguist-driven three-layered formalism, developed by UNDL Foundation. It generates natural language sentences out of semantic networks represented in the UNL format. In its current release, it is a web application developed in Java [1].

## 2. RELATED WORK

Initially '*DeCo*' was a tool for NLization, which was proposed by UNDL foundation in 2000. This DeConverter actually transforms the input, a UNL expression, into a directed graph structure with hyper-nodes. DeConversion of a UNL expression is carried out by applying DeConversion rules to these nodes. It processes the nodes from the entry node, to find an appropriate word for each node and generates a word sequence of target language. In this process, the system makes use of syntactic rules in order to produce syntactic structure of the target language and morphological rules to generate its morphemes, as depicted in Figure 2. DeConversion process ends when all words for all nodes are found and a word sequence of target sentence is completed.

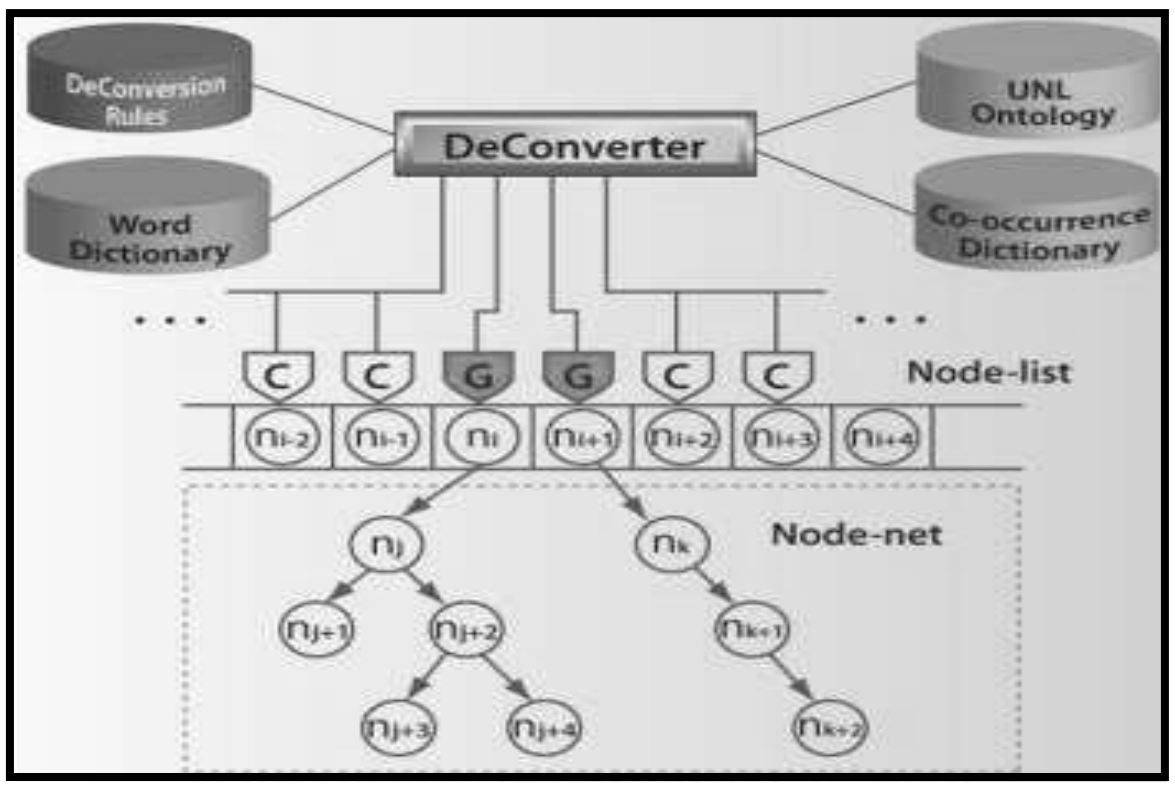

Figure 2. Structure of the DeConverter [2]

Here, *'G'* indicates generation windows and *'C'* indicates condition windows of the DeConverter. The DeConverter operates on the node-list through generation windows. Condition windows are used to check conditions for applying a rule. In the initial stage, the entry node of a UNL expression exists in the node-list and at the end of DeConversion, the node-list is converted to list of equivalent morphemes of the target language arranged syntactically to produce target language sentence [2].

## 3. EUGENE *SPECIFICATIONS*

EUGENE is a fully automatic engine. It takes a UNL input and delivers an output in natural language without any human intervention [3]. It is language-independent and as a universal engine, EUGENE must be parameterized to the target languages with few files for UNL-NL generation. These files are processed by EUGENE, through following tabs.

### 3.1. UNL Input

UNL Documents containing universal semantic network to be used for the generation of natural language texts are inputted into the system using first tab of EUGENE.

### 3.2. Dictionary

An editor and for inputting a lexical database, where root words are mapped into natural language entries, along with the corresponding features, to be provided according to the UNL Dictionary Specifications.

### 3.3. T-Rules

An editor and inputting transformation grammar(s), *i.e.*, a set of transformation rules used to convert the UNL graphs into natural language sentences, to be provided according to the UNL Grammar Specifications.

### 3.4. D-Rules

An editor and for inputting disambiguation grammar(s), *i.e.*, a set of disambiguation rules used to improve the results of the tokenization and of the transformation.

### 3.5. EUGENE

A console, from where the output corresponding to the UNL input sentence can be generated. The out can be displayed in five different trace levels. It takes a UNL input, processes it using T-rules and Dictionary loaded, finally, delivers an output in natural language without any human intervention.

In the next subsection features of Punjabi language and positioning of different part of speech in a sentence of Punjabi language discussed in brief.

## 4. FUTURE OF PUNJABI LANGUAGE

Punjabi is an Indo-Aryan language and is constitutionally recognized language of India. Indo-Aryan languages form a subgroup of the Indo-Iranian group of languages, which in turn belong to Indo-European family of languages. Punjabi is widely spoken in north-west India, Pakistan, United States, Australia, United Kingdom and Canada. There are more than 91 million native speakers of Punjabi language, which makes it approximately the 12th most widely spoken language in the world [4].

Punjabi has word classes in the form of noun, pronoun, adjective, cardinal, ordinal, main verb, auxiliary verb, adverb, postposition, conjunction, interjection and particle. Punjabi nouns change forms for number (singular or plural) and case in sentences. Punjabi nouns have assigned gender (masculine or feminine), *e.g.,* ਕੰਧ *kandh 'fence'* is used in feminine gender and ਮੇਜ਼ *mēz 'table'* is used in masculine gender. Punjabi has six types of pronouns. These are: personal pronouns, *e.g.,* ਮੈਂ maiṃ *'i'*; reflexive pronouns, *e.g.,* ਆਪ *āp* (some what equivalent to honorific form of English second person *'you'*); demonstrative pronouns, *e.g.,* ਉਹ *uh 'that'*; indefinite pronouns, *e.g.,* ਕੋਈ *kōī etc.*; relative pronouns (to join two clauses in a complex sentence), *e.g.,* ਜੋ *jō* and interrogative pronouns, *e.g.,* ਕੌਣ *kauṇ who'etc.* In Punjabi language, adjectives usually precede the nouns but follow the pronouns. The example of adjective preceding noun is, 'ਸੋਹਣਾ ਮੁੰਡਾ' *'sōhṇā muṇḍā' 'handsome boy'* and the example of adjective following pronoun is, 'ਮੈਂ ਸੋਹਣਾ ਹਾਂ ।' *'maiṃ sōhṇā hāṃ.' 'I am handsome'*. In a Punjabi sentence, verbs must agree with the subject or object of the sentence in terms of gender, number, and person. Punjabi verbs change forms for gender, number, person and tense. The verbs have assigned transitivity and causality. In Punjabi, there are two auxiliary verbs – ਹੈ *'hai'* for present tense (*e.g.,* ਰਾਮ ਅੰਬ ਖਾਂਦਾ ਹੈ । *'rām amb khāndā hai.' 'Ram eats mango'*) and ਸੀ *'sī'* for past tense (*e.g.,* ਰਾਮ ਨੇ ਅੰਬ ਖਾਧਾ ਸੀ । *'rām nē amb khādhā sī.' 'Ram had eaten mango'*). All the forms of these two auxiliary verbs can equally be used for both the genders [2]. For future tense in sentences, '*EGA'* form of main verb is used and in those sentences auxiliary.

## 5. IMPLEMENTATION FOR NLIZATION OF PUNJABI

The proposed NLization system for Punjabi has been tested on five types of part of speech sentences, which are, Determiners, Nouns, Verbs, Pronouns, Sentence Structure and Cardinals. Here, in this paper, NLization for Determiners, Verbs and Pronouns types of parts of speech, are discussed. Table 1 depicts the number T-rules and Dictionary words written corresponding to each type of part of speech.

Table1. Number of T-rules and Dictionary words for each type of part of speech sentences

| **Type** | **No. of Sentences** | **Rules** | **Dictionary Words** |
|---|---|---|---|
| Verbs | 50 | 25 | 3 |
| Pronouns | 40 | 8 | 20 |
| Determiners | 60 | 50 | 15 |

Detailed description for processing of these sentences is given in next subsection.

### 5.1. NLization of Verbs

The process of NLization of input UNL sentence containing verb to natural language sentence is illustrated with an example sentence given in (1).

Example English sentence is: He has arrived …(1)
UNL expression for this example sentence is given in (2).

```
{unl}
agt(arrive:0B.@present.@perfect., 00:01.@3.@male)
{/unl}                                                   …(2)
```

Equivalent Punjabi sentence: ਉਹ ਪਹੁੰਚ ਚੁੱਕਾ ਹੈ …(3)

Transliterated Punjabi sentence: *uh phunch chukka hai.*

UNL expression in this example contains two root words, first is a verb, *i.e., 'arrive'* and second is pronoun, *i.e., '00:01.@3'* /third person pronoun. The NLization process of UNL sentence is described in Table 2.

## 5.2. NLization of Pronouns

The process of NLization of input UNL sentence containing pronouns to natural language sentence is illustrated with an example sentence given in (4).

Example English sentence is: They love each other …(4)
UNL expression for this example sentence is given in (5).
{unl}
agt(love:03.@present.@reciprocal, 00:01.@3.@pl)
{/unl} …(5)
Equivalent Punjabi sentence is:
ਉਹ ਇਕ ਦੂਜੇ ਨੂੰ ਪਿਆਰ ਕਰਦੇ ਹਨ …(6)

Transliterated Punjabi sentence is: *uh ik duje nun pyar karde han.*

UNL expression in this example contains two root words, first is a verb, *i.e., 'arrive'* and second is pronoun, *i.e., '00:01.@3'* / third person pronoun.The NLization process UNL sentence is described in Table 3.

## 5.2. NLization of Determiners

The process of NLization of input UNL sentence containing determiners to natural language sentence is illustrated with an example sentence given in (7).

Example English sentence is: A lot of their books …(7)
UNL expression for this example sentence is given in (8).
{unl}
pos(book:05.@multal, 00:03.@3.@pl)
{/unl} …(8)
Equivalent Punjabi sentence is: ਉਹਨਾਂ ਦੀਆਂ ਬਹੁਤ ਕਿਤਾਬਾਂ …(9)

Transliterated Punjabi sentence is *uhna dian bhot kitaban*

UNL expression in this example contains two root words, first is a noun, *i.e., 'book'* and second is pronoun, *i.e., '00:01.@3'* or third person pronoun.The NLization process of example (10) UNL sentence is described in Table 4.

Table 2. NLization process of example (2) UNL sentence

| Rule Fired | Description | Action Taken |
|---|---|---|
| (%x,M7):=(%x,-M7,+FLX( PST&SNG&MCL&^PGS&^ANT:=0>" ਗਿਆ ਸੀ"; PST&SNG&FEM&^PGS&^ANT:=0>" ਗਈ ਸੀ"; PST&PLR&FEM&^ANT:=0>" ਗਈਆਂ ਸਨ"; | This paradigm M7 has been defined to attach corresponding postfix to verbs in Punjabi. Here attribute *'FLX'* indicate that word has been processed for inflection. | **To:** [arrive:07.@not.@present.@perfect] **Result:** ["ਪਹੁੰਚ":07.@not.@present.@perfect] Nothing is appended to UW |
| PST&PLR&MCL:=0>" ਗਏ ਸਨ";<br>PRS&SNG&MCL&^PGS&^PER:=0>"ਦਾ ਹੈ";<br>PRS&SNG&FEM&^PGS&^PER:=0>"ਦੀ ਹੈ";<br>PRS&PLR&FEM:=0>" ਦਿਆ ਹਨ"; PRS&PLR&MCL:=0>" ਦੇ ਹਨ";<br>PRS&SNG&FEM&^PGS&^PER:=0>"ਦੀ ਹੈ";<br>PRS&PLR&FEM:=0>" ਦਿਆ ਹਨ";<br>PRS&PLR&MCL:=0>" ਦੇ ਹਨ";<br>PRS&SNG&MCL&PGS:=0>" ਰਿਹਾ ਹੈ";<br>PRS&SNG&FEM&PGS:=0>" ਰਹੀ ਹੈ";<br>PRS&PLR&FEM&PGS:=0>" ਰਹੀਆਂ ਹਨ";<br>PRS&PLR&MCL&PGS:=0>" ਰਹੇ ਹਨ";<br>PST&SNG&MCL&PGS:=0>" ਰਿਹਾ ਸੀ";<br>PST&SNG&FEM&PGS:=0>" ਰਹੀ ਸੀ";<br>PST&PLR&FEM&PGS:=0>" ਰਹੀਆਂ ਸਨ";<br>PST&PLR&MCL&PGS:=0>" ਰਹੇ ਸਨ";<br>{PST&MCL&SNG&ANT}:=0>" ਚੁੱਕਾ ਸੀ";<br>{PST&FEM&SNG&ANT}:=0> "ਚੁੱਕੀ ਸੀ";<br>{PST&MCL&PLR&ANT}:=0>" ਚੁੱਕੇ ਸਨ";<br>{PST&FEM&PLR&ANT}:=0>" ਚੁੱਕੀਆਂ ਸਨ";<br>{PER&PRS&MCL&SNG}:=0> " | | 'ਪਹੁੰਚ' *phunch*, because this root word does not have any combination of attributes as are involved in this rule. |
| ਚੁੱਕਾ ਹੈ"; {PER&PRS&FEM&SNG}:=0>" ਚੁੱਕੀ ਹੈ"; {PER&FUT&MCL&SNG}:=0>" ਚੁੱਕਾ ਹੋਵੇਗਾ"; {PER&FUT&FEM&SNG}:=0>" ਚੁੱਕੀ ਹੋਵੇਗੀ"; {FUT&MCL&PGS&SNG}:=0>" ਰਿਹਾ ਹੋਵੇਗਾ"; {FUT&FEM&PGS&SNG}:=0>" ਰਿਹਾ ਹੋਵੇਗੀ";{FUT&MCL&SNG} :=0>"ੇਗਾ"; FUT&FEM&SNG&^PGS&^PER&^RES:=0>"ੇਗੀ";) ); | | |
| (%x,M2):=(%x,-M2,+FLX(AGT:=0>"ਨੂੰ")); | This paradigm M2 has been defined to attach word **"ਨੂੰ"** *nun* for UNL relation *'AGT'*. Here attribute *'FLX'* is indicate that word has been processed for inflections. | **To:** [00:01.@3.@male] **Result:**["ਉਹ":01.@3.@male] Nothing is appended to UW 'ਉਹ' *uh* because this root word does not have '*AGT'* attribute with it. |
| agt(%a,V,@present,@perfect,^PRS;%b,R):= agt(%a ,+PER,+PRS,-@present;%b); | It adds *'PER'* and '*PRS'* attributes to the verb depending upon attributes associated with the node, because *'PRS&PER'* combination is necessary for appending appropriate postfix to Punjabi root word verb. | **To:**agt(arrive:07.@not.@present.@perfect, 00:01.@3.@male) **Result:** agt(arrive:07.@not.@perfect, 00:01.@3) Nothing is appended to any Punjabi root word as postfix, *'@present'* attribute is removed from second node *'%b'* and *'PRS' and 'PES' are* added to first node *'%a'*. |
| agt(%a,^MCL;%b,R,@male):= agt(%a ,+MCL,-NUM,+NUM=SNG; %b,-@male); | It adds *'MCL'* and *'SNG'* attributes to the verb because of *'@male'* attribute associated with the second node '%b', as *'PRS&PER&MCL&SNG'* combination is necessary for appending appropriate postfix to Punjabi root word. | **To:**agt(arrive:07.@not.@present.@perfect, 00:01.@3.@male) **Result:** agt(arrive:07.@not.@perfect, 00:01.@3) Nothing is appended to any Punjabi root word as postfix, just *'@male'* attribute is removed from second node *'%b'* and *'MCL'* and *'SNG'* |

<table>
<tr><td></td><td></td><td>are added to first node ‘%a’.</td></tr>
<tr><td>agt(%a,V;%b,R):=(%b)(" ")(%a);</td><td>It resolves the ‘agt’ relation, places pronoun before verb and introduces new node of single space (“ ”) between two nodes.</td><td>To:<br>agt(arrive:07.@not.@perfect, 00:01.@3.@male)<br>Result:<br>#L(00:01.@3, -:02); #L(-:02, arrive:07.@not.@perfect)<br>Intermediate Output:<br>ਉਹ ਪਹੁੰਚ<br>uh phunch</td></tr>
<tr><td>({N|V|D|J|R},FLX,^inflected,%x):=(!FLX,-FLX,+inflected,%x);</td><td>It fires the corresponding paradigm rule to inflect the root word “ਉਹ” uh.</td><td>To:[00:01.@3]<br>Result:["ਉਹ":01.@3]<br>Nothing is appended to root word ‘ਉਹ’ uh because ‘AGT’ was not associated with it.</td></tr>
<tr><td>{N|V|D|J|R},FLX,^inflected,%x):=(!FLX,-FLX,+inflected,%x);</td><td>It fires the corresponding paradigm rule to inflect the root word “ਪਹੁੰਚ” phunch to make it “ਪਹੁੰਚ ਚੁੱਕਾ ਹੈ" phunch chukka hai</td><td>To:[arrive:07.@perfect]<br>Result:["ਪਹੁੰਚ ਚੁੱਕਾ ਹੈ" :07.@perfect]<br>Final Output:<br>ਉਹ ਪਹੁੰਚ ਚੁੱਕਾ ਹੈ<br>uh phunch chukka hai</td></tr>
</table>

Table 3. NLization process of example (5) UNL sentence

<table>
<tr><th>Rule Fired</th><th>Description</th><th>Action Taken</th></tr>
<tr><td>(%x,M2):=(%x,-M2,+FLX(AGT:=0>"ਨੂੰ"));</td><td>This paradigm M2 has been defined to attach word "ਨੂੰ" nun for UNL relation ‘AGT’. Here attribute ‘FLX’ is indicate that word has been processed for inflections.</td><td>To: [00:01.@3.@pl]<br>Result:["ਉਹ":01.@3.@pl]<br>Nothing is appended to root word ‘ਉਹ’ uh, because this root word does not have ‘AGT’ attribute with it.</td></tr>
<tr><td>agt(%a,V,@reciprocal;%b,@3,@pl):=(%b)<br>(" ")(%a,PER=3PS,+PLR);</td><td>It resolves ‘agt’ relationship, places pronoun before verb and introduces new node of single space (“ ”) between two nodes. It also adds ‘3PS’ and ‘PLR’ as attributes of verb, when attribute ‘@reciprocal’ was with first and ‘@3’, ‘@pl’ were associated with second node ‘%b’ in UNL sentence.</td><td>To:<br>agt(love:03.@present.@reciprocal, 00:01.@3.@pl)<br>Result:<br>#L(00:01.@3.@pl, -:02);<br>#L(-:02, love:03.@present.@reciprocal)<br>Intermediate Output:<br>ਉਹ ਪਿਆਰ<br>uh piyar</td></tr>
</table>

| | | |
|---|---|---|
| (V,@reciprocal,%x):=<br>("ਇਕ ਦੂਜੇ ਨੂੰ ")(%x,-@reciprocal); | It resolves attribute '*@reciprocal*' associated with verb and removes it from the verb after resolution. It adds "ਇਕ ਦੂਜੇ ਨੂੰ " *ik duje nun* before verb. | **To:**<br>[love:03.@present.@reciprocal]<br>**Result:**<br>#L(ਇਕ ਦੂਜੇ ਨੂੰ:04, -:05);<br>#L(-:05, love:03.@present)<br>**Intermediate Output:**<br>ਉਹ ਇਕ ਦੂਜੇ ਨੂੰ ਪਿਆਰ<br>*uh ik duje nu piyar* |
| (V,@present,ATE=INF,PER=3PS,NUM =PLR,%x):=(%x,-@present)(" ਕਰਦੇ ਹਨ"); | It appends " ਕਰਦੇ ਹਨ" *karde han* on right of verb by resolving its *'@present'* attribute , only if *'INF'* (indefinite tense) and 3PS, PLR (added manually by second fired rule) are associated with the verb. | **To:** [love:03.@present]<br>**Result:**<br>#L(love:03, -:06);<br>#L(-:06, ਕਰਦੇ ਹਨ:07)<br>**Intermediate Output:**<br>ਉਹ ਇਕ ਦੂਜੇ ਨੂੰ ਪਿਆਰ ਕਰਦੇ ਹਨ<br>*uh ik duje nu piyar* |
| ({N\|V\|D\|J\|R},FLX,^inflected,%x):=(!FLX,-FLX,+inflected,%x); | It fires the corresponding paradigm rule to inflect the root word "ਉਹ" *uh.* | **To:**[00:01.@3]<br>**Result:**["ਉਹ":01.@3]<br>**Final Output:**<br>ਉਹ ਇਕ ਦੂਜੇ ਨੂੰ ਪਿਆਰ ਕਰਦੇ ਹਨ<br>*uh ik duje nu piyar*<br>Nothing is appended to root word 'ਉਹ' *uh* because '*AGT'* was not associated with it. |

Table 4. NLization process of example (8) UNL sentence

| **Rule Fired** | **Description** | **Action Taken** |
|---|---|---|
| (%x,M2):=(%x,-M2,+FLX( SNG:=0>""; PLR:=0>"ਨਾਂ")); | This paradigm M2 has been defined to attach word "ਨਾਂ" *nan* when attribute *'PLR'* is associated with root word. Here attribute *'FLX'* is indicate that word has been processed for inflections. | **To:** [00:01.@3.@pl]<br>**Result:**["ਉਹ":01.@3.@pl]<br>Nothing is appended to root word 'ਉਹ' *uh* because this root word does not have '*PLR'* attribute with it. |
| (%x,M3):=(%x,M3,+FLX( SNG:=0>""; PLR:=0>"ਾਂ";)); | This paradigm M2 has been defined to attach word "ਾਂ" (to the right, for denoting multiple books in Punjabi) to the root word in Punjabi only | **To:** [book:05]<br>**Result:** ["ਕਿਤਾਬ":05]<br>Nothing is added because number information of the root word is *'SNG'* initially. |
| | when root word has *'PLR'* attribute associated with it. | |
| pos(%a,FEM,N,@multal;%b,@3,@pl,POD):=<br>(%b,+PLR)(" ਦੀਆਂ ") (%a); | It resolves the *'pos'* relationship, places pronoun before object, also introduces " ਦੀਆਂ " in b/w them because first node *'%a'* has attributes *'@multal'* (multiple) and *'FEM'* (female). | **To:** pos(book:05.@multal, 00:03.@3.@pl)<br>**Result:**<br>#L(00:03.@3.@pl, ਦੀਆਂ :01);<br>#L( ਦੀਆਂ :01,<br>book:05.@multal)<br>**Intermediate Output:**<br>ਉਹ ਦੀਆਂ ਕਿਤਾਬ<br>*uh dian kitab* |
| (N,@multal,%a):=<br>("ਬਹੁਤ ")<br>(%a,-@multal,-NUM, +NUM= PLR); | It resolves *'@multal'* (multiple) attribute by adding "ਬਹੁਤ " *boht* before the noun and updates its number information to *'PLR'* , *i.e.,* plural. | **To:** [book:05.@multal]<br>**Result:**<br>#L(ਬਹੁਤ:02, -:04); #L(-:04, book:05)<br>**Intermediate Output:**<br>ਉਹ ਦੀਆਂ ਬਹੁਤ ਕਿਤਾਬ<br>*uh dian boht kitab* |
| ({N\|V\|D},FLX,^inflected,%x):=(!FLX,-FLX,+inflected,%x); | It fires the corresponding paradigm rule to inflect the root word "ਕਿਤਾਬ" *kitab,* it attaches word "ਾਂ" (to the right, for denoting multiple books in Punjabi). | **To:** [book:05]<br>**Result:** ["ਕਿਤਾਬਾਂ":05]<br>**Intermediate Output:**<br>ਉਹ ਦੀਆਂ ਬਹੁਤ ਕਿਤਾਬਾਂ<br>*uh dian boht kitaban* |
| ({N\|V\|D},FLX,^inflected,%x):=(!FLX,-FLX,+inflected,%x); | It fires the corresponding paradigm rule to inflect the root word "ਉਹ" *uh*. It attaches word "ਨਾਂ" *nan* to the right, in Punjabi. | **To:** :["ਉਹ":01.@3.@pl]<br>**Result:** ["ਉਹਨਾਂ ":05]<br>**Final Output:**<br>ਉਹਨਾਂ ਦੀਆਂ ਬਹੁਤ ਕਿਤਾਬਾਂ<br>*uhna dian boht kitaban* |

## 6. RESULT AND DISCUSSION

In this paper, it is described in detail that, how UNL input sentence can be converted into Punjabi natural language, for Determiners, Verbs and Pronouns phrases using dictionary words and T-rules in the EUGENE console [5]. As a whole near about 140 sentences are processed successfully with exact desired output. All other sentences which fall in these three categories can also be processed easily by same set of written rules and by appending new dictionary words to the respective dictionaries.

There is a feature available on the UNDL foundation's website, which is **F-measure** [6], which rates the output of NLization (output generated using EUGENE tool) with the actual expected Punjabi natural langauge sentences on the scale of (0-1). We can upload both the files saved in UTF-8 .txt format and this gives us F-measure for it. F-measure for all the processed sentences is come out to be more than 90% as depicted in Figure 3.

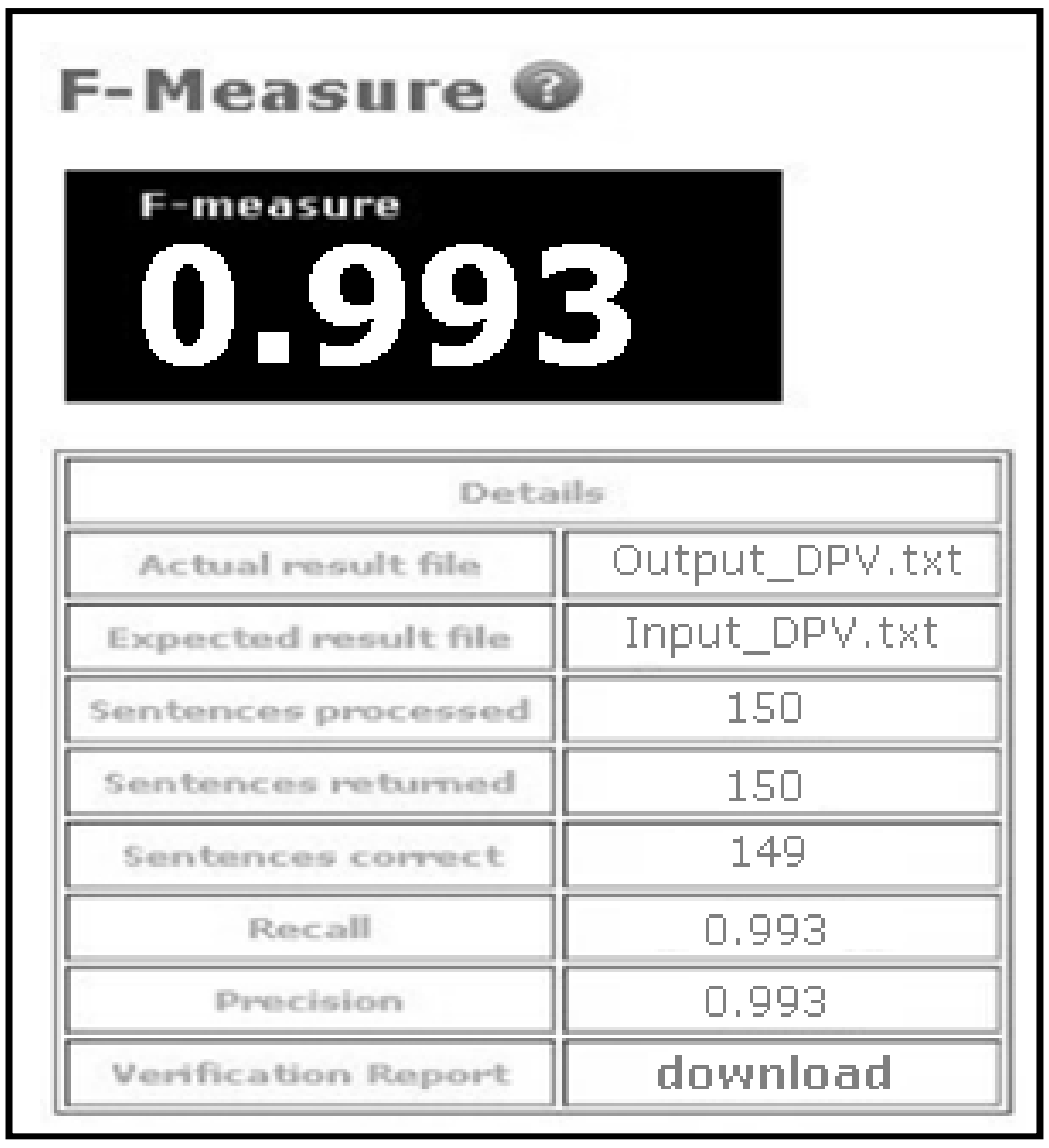

Figure 3. Combined F-measure for determiners, pronouns and verbs [7]

## 7. CONCLUSIONS

Many issues come while processing these UNL sentences, some changes made in the rules for the later sentences may cause the earlier sentences to get effected. Every time care should be taken while making changes to the rules for creating generalized rule sets. So writing the rules and editing them according to the grammar and sentence structure in Punjabi natural language is bit tedious task, but once the rules and dictionary gets written, we can process all other sentences which fall in these categories. EUGENE tool is very effective tool for accomplishing this task. It has very high quality because of its quality features like high availability, usability, high performance, due to these features, changes and crosschecking of sentences can be made readily.

## ACKNOWLEDGEMENTS

We would like to thank our colleague research scholars for their comments and suggestions for the paper. We would also like to thank our head of department Dr. Maninder Singh for motivating and for being ideal for us.

**Authors**

Harinder Singh is pursuing his Master of Engineering in Software Engineering from Thapar University, Patiala. He is undergoing the thesis of his M.E. entitled "Implementation of NLization Framework for Punjabi Language with EUGENE" in the supervision of Parteek Kumar. His current research interest includes UNL Based DeConversion, UNL Dictionary Specs and UNL Grammar Specs.

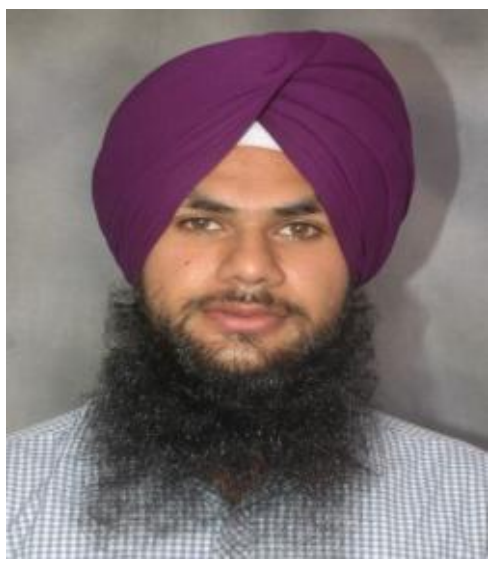

Parteek Kumar is Assistant Professor in the Department of Computer Science and Engineering at Thapar University, Patiala. He has more than fifteen years of academic experience. He has earned his B.Tech from SLIET and MS from BITS Pilani. He has completed his Ph.D on "UNL Based Machine Translation System for Punjabi Language" from Thapar University. He has published more than 50 research papers and articles in Journals, Conferences and Magazines of repute. His research work with UNDL foundation, Geneva, Switzerland provides international exposure to him and he was a member for Advanced UNL School at Alexandria, EGYPT in 2012. He has also undergone various faculty development programme from industries like Sun Microsystems, TCS and Infosys. He has authored six books including Simplified Approach to DBMS. He is acting as Co-PI for the research Project on Development of Punjabi WordNet sponsored By Department of Information Technology, Ministry of Communication and Information Technology, Govt. of India.

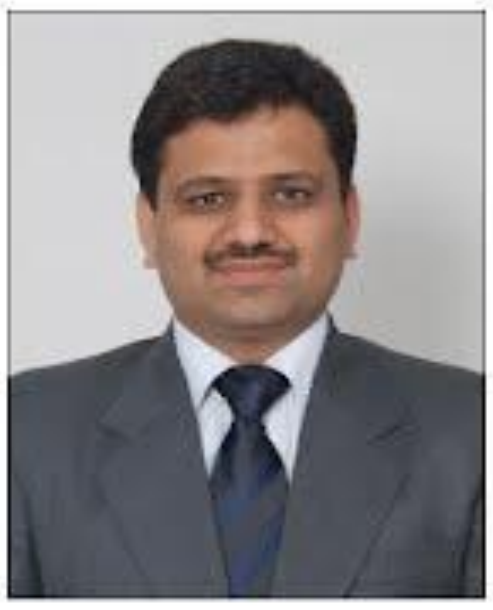